# Total-Order and Partial-Order Planning:
# A Comparative Analysis

**Steven Minton**                    MINTON@PTOLEMY.ARC.NASA.GOV
**John Bresina**                     BRESINA@PTOLEMY.ARC.NASA.GOV
**Mark Drummond**                    MED@PTOLEMY.ARC.NASA.GOV
*Recom Technologies*
*NASA Ames Research Center, Mail Stop: 269-2*
*Moffett Field, CA 94035 USA*

## Abstract

For many years, the intuitions underlying partial-order planning were largely taken for granted. Only in the past few years has there been renewed interest in the fundamental principles underlying this paradigm. In this paper, we present a rigorous comparative analysis of partial-order and total-order planning by focusing on two specific planners that can be directly compared. We show that there are some subtle assumptions that underlly the wide-spread intuitions regarding the supposed efficiency of partial-order planning. For instance, the superiority of partial-order planning can depend critically upon the search strategy and the structure of the search space. Understanding the underlying assumptions is crucial for constructing efficient planners.

## 1. Introduction

For many years, the superiority of partial-order planners over total-order planners has been tacitly assumed by the planning community. Originally, partial-order planning was introduced by Sacerdoti (1975) as a way to improve planning efficiency by avoiding "premature commitments to a particular order for achieving subgoals". The utility of partial-order planning was demonstrated anecdotally by showing how such a planner could efficiently solve blocksworld examples, such as the well-known "Sussman anomaly".

Since partial-order planning intuitively seems like a good idea, little attention has been devoted to analyzing its utility, at least until recently (Minton, Bresina, & Drummond, 1991a; Barrett & Weld, 1994; Kambhampati, 1994c). However, if one looks closely at the issues involved, a number of questions arise. For example, do the advantages of partial-order planning hold regardless of the search strategy used? Do the advantages hold when the planning language is so expressive that reasoning about partially ordered plans is intractable (e.g., if the language allows conditional effects)?

Our work (Minton et al., 1991a, 1992) has shown that the situation is much more interesting than might be expected. We have found that there are some "unstated assumptions" underlying the supposed efficiency of partial-order planning. For instance, the superiority of partial-order planning *can* depend critically upon the search strategy and search heuristics employed.

This paper summarizes our observations regarding partial-order and total-order planning. We begin by considering a simple total-order planner and a closely related partial-order planner and establishing a mapping between their search spaces. We then examine the relative sizes of their search spaces, demonstrating that the partial-order planner has a fundamental advantage because the size of its search space is always less than or equal to that of the total-order planner. However, this advantage does not necessarily translate into an efficiency gain; this depends on the type of search strategy used. For example, we describe a domain where our partial order planner is more efficient than our total order planner when depth-first search is used, but the efficiency gain is lost when an iterative sampling strategy is used.

We also show that partial-order planners can have a second, independent advantage when certain types of operator ordering heuristics are employed. This "heuristic advantage" underlies Sacerdoti's anecdotal examples explaining why least-commitment works. However, in our blocksworld experiments, this second advantage is relatively unimportant compared to the advantage derived from the reduction in search space size.

Finally, we look at how our results extend to partial-order planners in general. We describe how the advantages of partial-order planning can be preserved even if highly expressive languages are used. We also show that the advantages do not *necessarily* hold for all partial-order planners, but depend critically on the construction of the planning space.

## 2. Background

Planning can be characterized as search through a space of possible plans. A *total-order planner* searches through a space of totally ordered plans; a *partial-order planner* is defined analogously. We use these terms, rather than the terms "linear" and "nonlinear", because the latter are overloaded. For example, some authors have used the term "nonlinear" when focusing on the issue of *goal ordering*. That is, some "linear" planners, when solving a conjunctive goal, require that all subgoals of one conjunct be achieved before subgoals of the others; hence, planners that can arbitrarily interleave subgoals are often called "nonlinear". This version of the linear/nonlinear distinction is different than the partial-order/total-order distinction investigated here. The former distinction impacts planner completeness, whereas the total-order/partial-order distinction is orthogonal to this issue (Drummond & Currie, 1989; Minton et al., 1991a).

The total-order/partial-order distinction should also be kept separate from the distinction between "world-based planners" and "plan-based planners". The distinction is one of modeling: in a world-based planner, each search state corresponds to a state of the world and in a plan-based planner, each search state corresponds to a plan. While total-order planners are commonly associated with world-based planners, such as STRIPS, several well-known total-order planners have been plan-based, such as Waldinger's regression planner (Waldinger, 1975), Interplan (Tate, 1974) and Warplan (Warren, 1974). Similarly, partial-order planners are commonly plan-based, but it is possible to have a world-based partial-order planner (Godefroid & Kabanza, 1991). In this paper, we focus solely on the total-order/partial-order distinction in order to avoid complicating the analysis.

We claim that the only significant difference between partial-order and total-order planners is planning efficiency. It might be argued that partial-order planning is preferable because a partially ordered plan can be more flexibly executed. However, execution flexibility can also be achieved with a total-order planner and a post-processing step that removes unnecessary orderings from the totally ordered solution plan to yield a partial order (Back-strom, 1993; Veloso, Perez, & Carbonell, 1990; Regnier & Fade, 1991). The polynomial time complexity of this post-processing is negligible compared to the search time for plan generation.[1] Hence, we believe that execution flexibility is, at best, a weak justification for the supposed superiority of partial-order planning.

In the following sections, we analyze the relative efficiency of partial-order and total-order planning by considering a total-order planner and a partial-order planner that can be directly compared. Elucidating the key differences between these planning algorithms reveals some important principles that are of general relevance.

## 3. Terminology

A plan consists of an ordered set of *steps*, where each step is a unique operator instance. Plans can be *totally ordered*, in which case every step is ordered with respect to every other step, or *partially ordered*, in which case steps can be unordered with respect to each other. We assume that a library of operators is available, where each operator has preconditions, deleted conditions, and added conditions. All of these conditions must be nonnegated propositions, and we adopt the common convention that each deleted condition is a precondition. Later in this paper we show how our results can be extended to more expressive languages, but this simple language is sufficient to establish the essence of our argument.

A *linearization* of a partially ordered plan is a total order over the plan's steps that is consistent with the existing partial order. In a totally ordered plan, a precondition of a plan step is *true* if it is added by an earlier step and not deleted by an intervening step. In a partially ordered plan, a step's precondition is *possibly true* if there exists a linearization in which it is true, and a step's precondition is *necessarily true* if it is true in *all* linearizations. A step's precondition is *necessarily false* if it is not possibly true.

A *state* consists of a set of propositions. A *planning problem* is defined by an *initial state* and a set of *goals*, where each goal is a proposition. For convenience, we represent a problem as a two-step *initial plan*, where the propositions that are true in the initial state are added by the first step, and the goal propositions are the preconditions of the final step. The planning process starts with this initial plan and searches through a space of possible plans. A successful search terminates with a *solution* plan, i.e., a plan in which all steps' preconditions are necessarily true. The search space can be characterized as a tree, where each node corresponds to a plan and each arc corresponds to a plan transformation. Each transformation incrementally extends (i.e., refines) a plan by adding additional steps or orderings. Thus, each leaf in the search tree corresponds either to a solution plan or a dead-end, and each intermediate node corresponds to an unfinished plan which can be further extended.

---

1. Backstrom (1993) formalizes the problem of removing unnecessary orderings in order to produce a "least-constrained" plan. He shows that the problem is polynomial if one defines a least-constrained plan as a plan in which no orderings can be removed without impacting the correctness of the plan. Backstrom also shows that the problem of finding a plan with the fewest orderings over a given operator set is a much harder problem; it is NP-hard.

TO($P, G$)

1. **Termination check:** If $G$ is empty, report success and return solution plan P.

2. **Goal selection:** Let $c$ = select-goal($G$), and let $O_{need}$ be the plan step for which $c$ is a precondition.

3. **Operator selection:** Let $O_{add}$ be an operator in the library that adds $c$. If there is no such $O_{add}$, then terminate and report failure. *Choice point: all such operators must be considered for completeness.*

4. **Ordering selection:** Let $O_{del}$ be the last deleter of $c$. Insert $O_{add}$ somewhere between $O_{del}$ and $O_{need}$, call the resulting plan $P'$. *Choice point: all such positions must be considered for completeness.*

5. **Goal updating:** Let $G'$ be the set of preconditions in $P'$ that are not true.

6. **Recursive invocation:** TO($P', G'$).

Figure 1: The TO planning algorithm

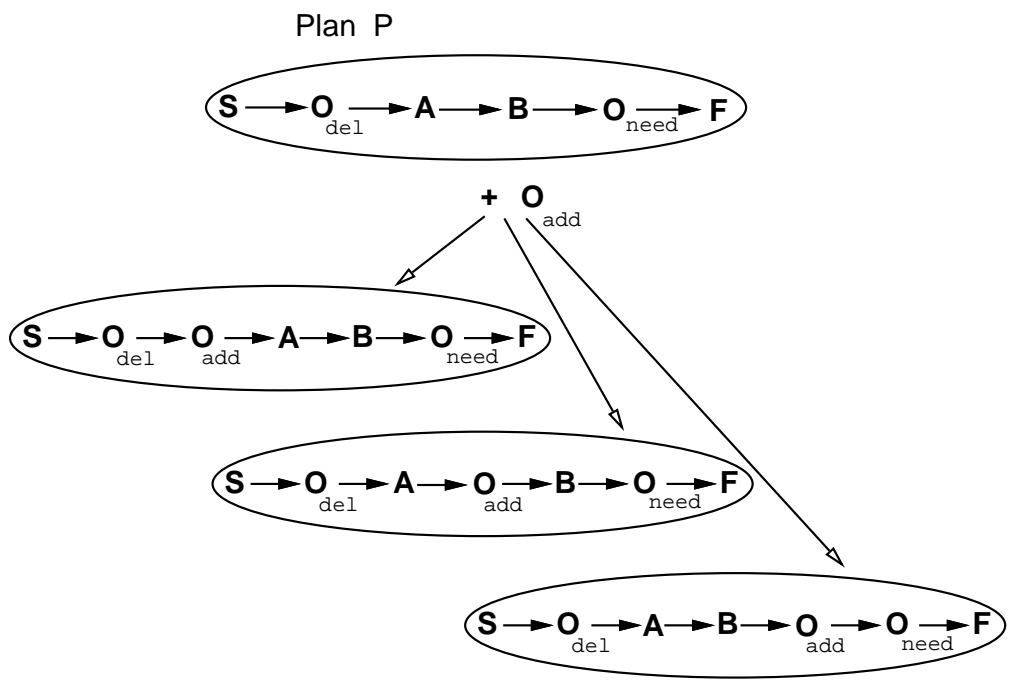

Figure 2: How TO extends a plan: Adding $O_{add}$ to plan $P$ generates three alternatives.

## 4. A Tale of Two Planners

In this section we define two simple planning algorithms. The first algorithm, shown in Figure 1, is TO, a total-order planner motivated by Waldinger's regression planner (Waldinger, 1975), Interplan (Tate, 1974), and Warplan (Waldinger, 1975). Our purpose here is to characterize the search space of the TO planning algorithm, and the pseudo-code in Figure 1 accomplishes this by defining a nondeterministic procedure that enumerates possible plans. (If the plans are enumerated by a breadth-first search, then the algorithms presented in this section are provably complete, as shown in Appendix A.)

TO accepts an unfinished plan, $P$, and a goal set, $G$, containing preconditions which are currently not true. If the algorithm terminates successfully then it returns a totally ordered solution plan. Note that there are two choice points in this procedure: operator selection and ordering selection. The procedure does not need to consider alternative goal choices. For our purposes, the function `select-goal` can be any deterministic function that selects a member of $G$.

As used in Step 4, the *last deleter* of a precondition $c$ for a step $O_{need}$ is defined as follows. Step $O_{del}$ is the last deleter of $c$ if $O_{del}$ deletes $c$, $O_{del}$ is before $O_{need}$, and there is no other deleter of $c$ between $O_{del}$ and $O_{need}$. In the case that no step before $O_{need}$ deletes $c$, the first step is considered to be the last deleter.

Figure 2 illustrates TO's plan extension process. This example assumes that steps $A$ and $B$ do not add or delete $c$. There are three possible insertion points for $O_{add}$ in plan $P$, each yielding an alternative extension.

The second planner is UA, a partial-order planner, shown in Figure 3. UA is similar to TO in that it uses the same procedures for goal selection and operator selection; however, the procedure for ordering selection is different. Step 4 of UA inserts orderings, but only "interacting" steps are ordered. Specifically, we say that two steps *interact* if they are unordered with respect to each other and either:

- one step has a precondition that is added or deleted by the other step, or

- one step adds a condition that is deleted by the other step.

The only significant difference between UA and TO lies in Step 4: TO orders the new step with respect to *all* others, whereas UA adds orderings only to eliminate interactions. It is in this sense that UA is *less committed* than TO.

Figure 4 illustrates UA's plan extension process. As in Figure 2, we assume that steps $A$ and $B$ do not add or delete $c$; however, step $A$ and $O_{add}$ interact with respect to some other condition. This interaction yields two alternative plan extensions: one in which $O_{add}$ is ordered before $A$ and one in which $O_{add}$ is ordered after $A$.

Since UA orders all steps which interact, the plans that are generated have a special property: each precondition in a plan is either *necessarily* true or *necessarily* false. We call such plans *unambiguous*. This property yields a tight correspondence between the two planners' search spaces. Suppose UA is given the unambiguous plan $U$ and TO is given the plan $T$, where $T$ is a linearization of $U$. Let us consider the relationship between the way that UA extends $U$ and TO extends $T$. Note that the two planners will have the same set of goals since, by definition, each goal in $U$ is a precondition that is necessarily false, and a precondition is necessarily false if and only if it is false in every linearization. Since the two plans have the same set of goals and since both planners use the same goal selection method, both algorithms pick the same goal; therefore, $O_{need}$ is the same for both. Similarly, both algorithms consider the same library operators to achieve this goal. Since $T$ is a linearization of $U$, and $O_{need}$ is the same in both plans, both algorithms find the same last deleter as well.[2] When TO adds a step to a plan, it orders the new step with respect to

---

2. There is a unique last deleter in $U$. This follows from our requirement that for any operator in our language, the deleted conditions must be a subset of the preconditions. If two unordered steps delete the same condition, then that condition must also be a precondition of both operators. Hence, the two steps interact and will be ordered by UA.

UA($P, G$)

1. **Termination check:** If $G$ is empty, report success and return solution plan P.

2. **Goal selection:** Let $c$ = select-goal($G$), and let $O_{need}$ be the plan step for which $c$ is a precondition.

3. **Operator selection:** Let $O_{add}$ be an operator in the library that adds $c$. If there is no such $O_{add}$, then terminate and report failure. *Choice point: all such operators must be considered for completeness.*

4. **Ordering selection:** Let $O_{del}$ be the last deleter of $c$. Order $O_{add}$ after $O_{del}$ and before $O_{need}$. Repeat until there are no interactions:
   ○ Select a step $O_{int}$ that interacts with $O_{add}$.
   ○ Order $O_{int}$ either before or after $O_{add}$.
   *Choice point: both orderings must be considered for completeness.*
   Let $P'$ be the resulting plan.

5. **Goal updating:** Let $G'$ be the set of preconditions in $P'$ that are necessarily false.

6. **Recursive invocation:** UA($P', G'$).

Figure 3: The UA planning algorithm

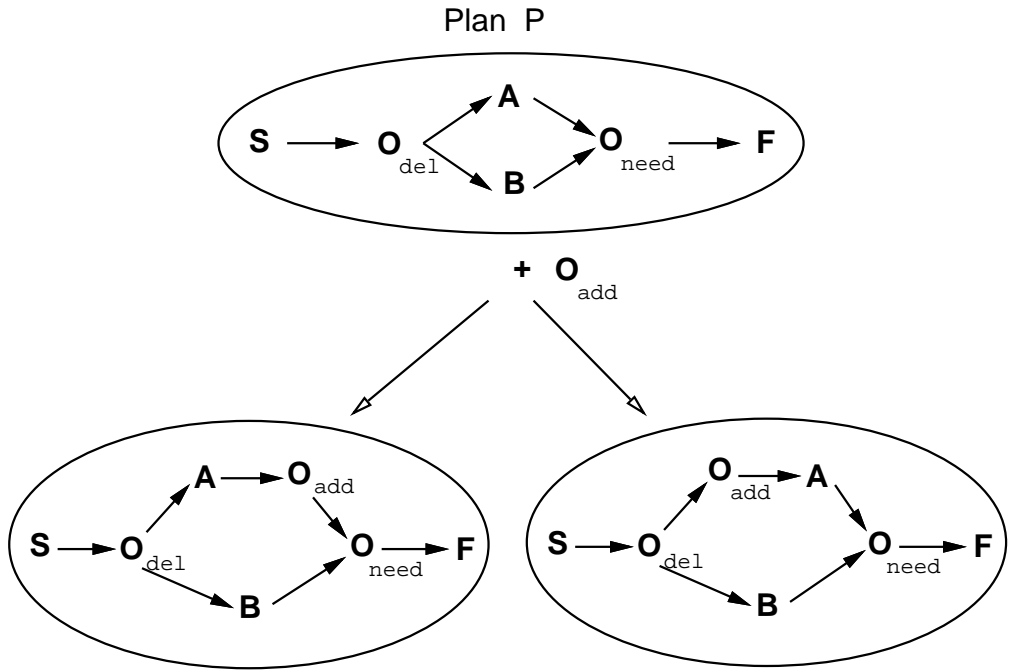

Figure 4: How UA extends a plan: Adding $O_{add}$ to plan $P$ generates two alternatives. The example assumes that $O_{add}$ interacts with step $A$.

all existing steps. When UA adds a step to a plan, it orders the new step *only* with respect to interacting steps. UA considers all possible combinations of orderings which eliminate interactions; hence, for any plan produced by TO, UA produces a corresponding plan that is less-ordered or equivalent.

The following sections exploit this tight correspondence between the search spaces of UA and TO. In the next section we analyze the relative sizes of the two planners' search spaces, and later we compare the number of plans actually generated under different search strategies.

## 5. Search Space Comparison

The search space for both TO and UA can be characterized as a tree of plans. The root node in the tree corresponds to the top-level invocation of the algorithm, and the remaining nodes each correspond to a recursive invocation of the algorithm. Note that in generating a plan, the algorithms make both operator and ordering choices, and each different set of choices corresponds to a single branch in the search tree.

We denote the search tree for TO by $tree_{TO}$ and, similarly, the search tree for UA by $tree_{UA}$. The number of plans in a search tree is equal to the number of times the planning procedure (UA or TO) would be invoked in an exhaustive exploration of the search space. Note that every plan in $tree_{UA}$ and $tree_{TO}$ is unique, since each step in a plan is given a unique label. Thus, although two plans in the same tree might both be instances of a particular operator sequence, such as $O1 \prec O2 \prec O3$, the plans are distinct because their steps have different labels. (We have defined our plans this way to make our proofs more concise.)

We can show that for any given problem, $tree_{TO}$ is at least as large as $tree_{UA}$, that is, the number of plans in $tree_{TO}$ is greater than or equal to the number of plans in $tree_{UA}$. This is done by proving the existence of a function $\mathcal{L}$ which maps plans in $tree_{UA}$ into sets of plans in $tree_{TO}$ that satisfies the following two conditions.

1. **Totality Property**: For every plan $U$ in $tree_{UA}$, there exists a non-empty set $\{T_1, \ldots, T_m\}$ of plans in $tree_{TO}$ such that $\mathcal{L}(U) = \{T_1, \ldots, T_m\}$.

2. **Disjointness Property**: $\mathcal{L}$ maps distinct plans in $tree_{UA}$ to disjoint sets of plans in $tree_{TO}$; that is, if $U_1, U_2 \in tree_{UA}$ and $U_1 \neq U_2$, then $\mathcal{L}(U_1) \cap \mathcal{L}(U_2) = \{\}$.

Let us examine why the existence of an $\mathcal{L}$ with these two properties is sufficient to prove that the size of UA's search tree is no greater than that of TO. Figure 5 provides a guide for the following discussion. Intuitively, we can use $\mathcal{L}$ to count plans in the two search trees. For each plan counted in $tree_{UA}$, we use $\mathcal{L}$ to count a non-empty set of plans in $tree_{TO}$. The totality property means that every time we count a plan in $tree_{UA}$, we count at least one plan in $tree_{TO}$; this implies that $\mid tree_{UA} \mid \ \leq \ \sum_{U \in tree_{UA}} \mid \mathcal{L}(U) \mid$. Of course, we must further show that each plan counted in $tree_{TO}$ is counted only once; this is guaranteed by the disjointness property, which implies that $\sum_{U \in tree_{UA}} \mid \mathcal{L}(U) \mid \ \leq \ \mid tree_{TO} \mid$. Thus, the conjunction of the two properties implies that $\mid tree_{UA} \mid \ \leq \ \mid tree_{TO} \mid$.

We can define a function $\mathcal{L}$ that has these two properties as follows. Let $U$ be a plan in $tree_{UA}$, let $T$ be a plan in $tree_{TO}$, and let *parent* be a function from a plan to its parent plan in the tree. Then $T \in \mathcal{L}(U)$ if and only if (*i*) $T$ is a linearization of $U$ and (*ii*) either $U$ and $T$ are both root nodes of their respective search trees or $parent(T) \in \mathcal{L}(parent(U))$. Intuitively, $\mathcal{L}$ maps a plan $U$ in $tree_{UA}$ to all linearizations which share common derivation ancestry.[3] This is illustrated in Figure 5, where for each plan in $tree_{UA}$ a dashed line is drawn to the corresponding set of plans in $tree_{TO}$.

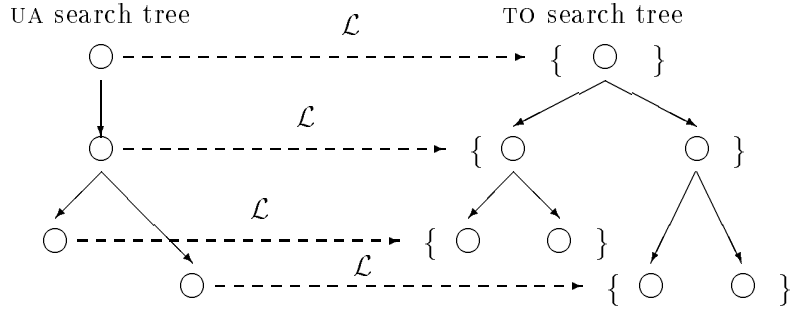

Figure 5: How $\mathcal{L}$ maps from $tree_{UA}$ to $tree_{TO}$

We can show that $\mathcal{L}$ satisfies the totality and disjointness properties by induction on the depth of the search trees. Detailed proofs are in the appendix. To prove the first property, we show that for every plan contained in $tree_{UA}$, all linearizations of that plan are contained in $tree_{TO}$. To prove the second property, we note that any two plans at different depths in $tree_{UA}$ have disjoint sets of linearizations, and then show by induction that any two plans at the same depth in $tree_{UA}$ also have this property.

How much smaller is $tree_{UA}$ than $tree_{TO}$? The mapping described above provides an answer. For each plan $U$ in $tree_{UA}$ there are $\mid \mathcal{L}(U) \mid$ distinct plans in TO, where $\mid \mathcal{L}(U) \mid$ is the number of linearizations of $U$. The exact number depends on how unordered $U$ is. A totally unordered plan has a factorial number of linearizations and a totally ordered plan has only a single linearization. Thus, the only time that the size of $tree_{UA}$ equals the size of $tree_{TO}$ is when every plan in $tree_{UA}$ is totally ordered; otherwise, $tree_{UA}$ is strictly smaller than $tree_{TO}$ and possibly exponentially smaller.

## 6. Time Cost Per Plan

While the size of UA's search tree is possibly exponentially smaller than that of TO, it does not follow that UA is necessarily more efficient. Efficiency is determined by two factors: the time cost per plan in the search tree (discussed in this section) and the size of the subtree explored during the search process (discussed in the next section).

---

3. The reader may question why $\mathcal{L}$ maps $U$ to all its linearizations in $tree_{TO}$ that share common derivation ancestry, as opposed to simply mapping $U$ to all its linearizations in $tree_{TO}$. The reason is that our planners are not systematic, in the sense that they may generate two or more plans with the same operator sequence. We can distinguish such plans by their derivational history. For example, suppose two instantiations of the same operator sequence $O1 \prec O2 \prec O3$ exist within a $tree_{TO}$ but they correspond to different plans in $tree_{UA}$. $\mathcal{L}$ relies on their different derivations to determine the appropriate correspondence.

| Step | Executions Per Plan | TO Cost | UA Cost |
|------|---------------------|---------|---------|
| 1 | 1 | $O(1)$ | $O(1)$ |
| 2 | 1 | $O(1)$ | $O(1)$ |
| 3 | $< 1$ | $O(1)$ | $O(1)$ |
| 4 | 1 | $O(1)$ | $O(e)$ |
| 5 | 1 | $O(n)$ | $O(e)$ |

Table 1: Cost per plan comparisons

In this section we show that while UA can indeed take more time per plan, the extra time is relatively small and grows only polynomially with the number of steps in the plan,[4] which we denote by $n$. In comparing the relative efficiency of UA and TO, we first consider the number of times that each algorithm step is executed per plan in the search tree and we then consider the time complexity of each step.

As noted in the preceding sections, each node in the search tree corresponds to a plan, and each invocation of the planning procedure for both UA and TO corresponds to an attempt to extend that plan. Thus, for both UA and TO, it is clear that the termination check and goal selection (Steps 1 and 2) are each executed once per plan. Analyzing the number of times that the remaining steps are executed might seem more complicated, since each of these steps is executed many times at an internal node and not at all at a leaf. However, the analysis is actually quite simple since we can amortize the number of executions of each step over the number of plans produced. Notice that Step 6 is executed once for each plan that is generated (i.e., once for each node other than the root node). This gives us a bound on the number of times that Steps 3, 4, and 5 are executed.[5] More specifically, for both algorithms, Step 3 is executed fewer times than Step 6, and Steps 4 and 5 are executed exactly the same number of times that Step 6 is executed, that is, once for each plan that is generated. Consequently, for both algorithms, no step is executed more than once per plan, as summarized in Table 1. In other words, the number of times each step is executed during the planning process is bounded by the size of the search tree.

In examining the costs for each step, we first note that for both algorithms, Step 1, the termination check, can be accomplished in $O(1)$ time. Step 2, goal selection, can also be accomplished in $O(1)$ time; for example, assuming the goals are stored in a list, the `select-goal` function can simply return the first member of the list. Each execution of Step 3, operator selection, also only requires $O(1)$ time; if we assume the operators are indexed by their effects, all that is required is to "pop" the list of relevant operators on each execution.

---

4. We assume that the size of the operators (the number of preconditions and effects) is bounded by a constant for a given domain.

5. Since Steps 3 and 4 are nondeterministic, we need to be clear about our terminology. We say that Step 3 is executed once each time a different operator is chosen, and Step 4 is executed once for each different combination of orderings that is selected.

Steps 4 and 5 are less expensive for TO than for UA. Step 4 of TO is accomplished by inserting the new operator, $O_{add}$, somewhere between $O_{del}$ and $O_{need}$. If the possible insertion points are considered starting at $O_{need}$ and working towards $O_{del}$, then each execution of Step 4 can be accomplished in constant time, since each insertion constitutes one execution of the step. In contrast, Step 4 in UA involves carrying out interaction detection and elimination in order to produce a new plan $P'$. This step can be accomplished in $O(e)$ time, where $e$ is the number of edges in the graph required to represent the partially ordered plan. (In the worst case, there may be $O(n^2)$ edges in the plan, and in the best case, $O(n)$ edges.) The following is the description of UA's ordering step, from Figure 3, with some additional implementation details:

4. **Ordering selection**: Order $O_{add}$ after $O_{del}$ and before $O_{need}$. Label all steps preceding $O_{add}$ and all steps following $O_{add}$. Let $steps_{int}$ be the unlabeled steps that interact with $O_{add}$. Let $O_{del}$ be the last deleter of $c$. Repeat until $steps_{int}$ is empty:

   - Let $O_{int} = \text{Pop}(steps_{int})$
   - if $O_{int}$ is still unlabeled then either:
     - order $O_{int}$ before $O_{add}$, and label $O_{int}$ and the unlabeled steps before $O_{int}$; or
     - order $O_{int}$ after $O_{add}$, and label $O_{int}$ and the unlabeled steps after $O_{int}$.

     *Choice point: both orderings must be considered for completeness.*

   Let $P'$ be the resulting plan.

The ordering process begins with a preprocessing stage. First, all steps preceding or following $O_{add}$ are labeled as such. The labeling process is implemented by a depth-first traversal of the plan graph, starting with $O_{add}$ as the root, which first follows the edges in one direction and then follows edges in the other direction. This requires at most $O(e)$ time. After the labeling process is complete, only steps that are unordered with respect to $O_{add}$ are unlabeled, and thus the interacting steps (which must be unordered with respect to $O_{add}$) are identifiable in $O(n)$ time. The last deleter is identifiable in $O(e)$ time.

After the preprocessing stage, the procedure orders each interacting step with respect to $O_{add}$, updating the labels after each iteration. Since each edge in the graph need be traversed no more than once, the entire ordering process takes at most $O(e)$ time (as described in Minton et al., 1991b). To see this, note that the process of labeling the steps before (or after) $O_{int}$ can stop as soon as a labeled step is encountered.

Having shown that Step 4 of TO has $O(1)$ complexity and Step 4 of UA has $O(e)$ complexity, we now consider Step 5 of both algorithms, updating the goal set. TO accomplishes this by iterating through the steps in the plan, from the head to the tail, which requires $O(n)$ time. UA accomplishes this in a similar manner, but it requires $O(e)$ time to traverse the graph. (Alternatively, UA can use the same procedure as TO, provided an $O(e)$ topological sort is first done to linearize the plan.)

To summarize our complexity analysis, the use of a partial order means that UA incurs greater cost for operator ordering (Step 4) and for updating the goal set (Step 5). Overall, UA requires $O(e)$ time per plan, while TO only requires $O(n)$ time per plan. Since a totally ordered plan requires a representation of size $O(n)$, and a partially ordered graph requires a representation of size $O(e)$, designing procedures with lower costs would be possible only if the entire plan graph did not need to be examined in the worst case.

## 7. The Role of Search Strategies

The previous sections have compared TO and UA in terms of relative search space size and relative time cost per node. The extra processing time required by UA for each node would appear to be justified since its search space may contain exponentially fewer nodes. However, to complete our analysis, we must consider the number of nodes actually visited by each algorithm under a given search strategy.

For breadth-first search, the analysis is straightforward. After completing the search to a particular depth, both planners will have explored their entire trees up to that depth.[6] Both UA and TO find a solution at the same depth due to the correspondence between their search trees. Thus, the degree to which UA will outperform TO, under breadth-first, depends solely on the "expansion factor" under $\mathcal{L}$, i.e., on the number of linearizations of UA's plans.

We can formalize this analysis as follows. For a node $U$ in $tree_{UA}$, we denote the number of steps in the plan at $U$ by $n_u$, and the number of edges in $U$ by $e_u$. Then for each node $U$ that UA generates, UA incurs time cost $O(e_u)$; whereas, TO incurs time cost $O(n_u) \cdot | \mathcal{L}(U) |$, where $| \mathcal{L}(U) |$ is the number of linearizations of the plan at node $U$. Therefore, the ratio of the total time costs of TO and UA is as follows, where $bf(tree_{UA})$ denotes the subtree considered by UA under breadth-first search.

$$\frac{\text{cost}(\text{TO}_{bf})}{\text{cost}(\text{UA}_{bf})} = \frac{\sum_{u \in bf(tree_{UA})} O(n_u) \cdot | \mathcal{L}(U) |}{\sum_{u \in bf(tree_{UA})} O(e_u)}$$

The analysis of breadth-first search is so simple because this search strategy preserves the correspondence between the two planners' search spaces. In breadth-first search, the two planners are *synchronized* after exhaustively exploring each level, so that TO has explored (exactly) the linearizations of the plans explored by UA. For any other search strategy which similarly preserves the correspondence, such as iterative deepening, a similar analysis can be carried out.

The cost comparison is not so clear-cut for depth-first search, since the correspondence is not guaranteed to be preserved. It is easy to see that, under depth-first search, TO does not necessarily explore all linearizations of the plans explored by UA. This is not simply because the planners nondeterministically choose which child to expand. There is a deeper reason: the correspondence $\mathcal{L}$ does not preserve the subtree structure of the search space. For a plan $U$ in $tree_{UA}$, the corresponding linearizations in $\mathcal{L}(U)$ may be spread throughout $tree_{TO}$. Therefore, it is unlikely that corresponding plans will be considered in the same order by depth-first search. Nevertheless, even though the two planners are not synchronized, we might expect that, on average, UA will explore fewer nodes because the size of $tree_{UA}$ is less than or equal to the size of $tree_{TO}$.

Empirically, we have observed that UA does tend to outperform TO under depth-first search, as illustrated by the experimental results in Figure 6. The first graph compares the mean number of *nodes* explored by UA and TO on 44 randomly generated blocksworld problems; the second graph compares the mean planning *time* for UA and TO on the same problems and demonstrates that the extra time cost per node for UA is relatively insignificant. The problems are partitioned into 4 sets of 11 problems each, according to minimal solution "length" (i.e., the number of steps in the plan). For each problem, both planners were given a depth-limit equal to the length of the shortest solution.[7] Since the planners make nondeterministic choices, 25 trials were conducted for each problem. The source code and data required to reproduce these experiments can be found in Online Appendix 1.

---

6. For perspicuity, we ignore the fact that the number of nodes explored by the two planners on the last level may differ if the planners stop when they reach the first solution.

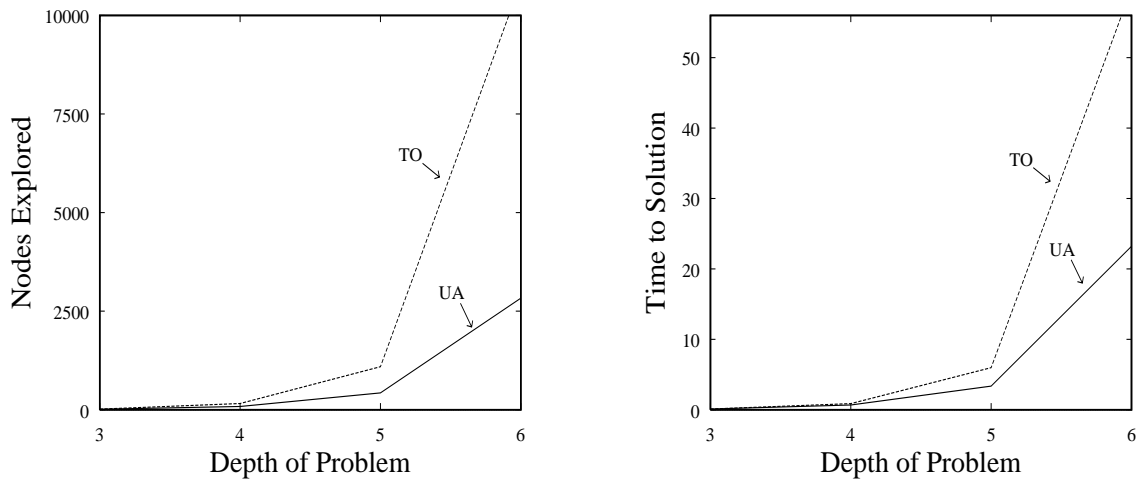

Figure 6: UA and TO Performance Comparison under Depth-First Search

As we pointed out, one plausible explanation for the observed dominance of UA is that TO's search tree is at least as large as UA's search tree. In fact, in the above experiments we often observed that TO's search tree was typically much larger. However, the full story is more interesting. Search tree size alone is not sufficient to explain UA's dominance; in particular, the density and distribution of solutions play an important role.

The solution density of a search tree is the proportion of nodes that are solutions.[8] If the solution density for TO's search tree is greater than that for UA's search tree, then TO might outperform UA under depth-first search *even though* TO's search tree is actually larger. For example, it might be the case that all UA solution plans are completely unordered and that the plans at the remaining leaves of $tree_{UA}$ – the *failed plans* – are totally ordered. In this case, each UA solution plan corresponds to an exponential number of TO solution plans, and each UA failed plan corresponds to a single TO failed plan. The converse is also possible: the solution density of UA's search tree might be greater than that of TO's search tree, thus favoring UA over TO under depth-first search. For example, there might be a single totally ordered solution plan in UA's search tree and a large number of highly unordered failed plans. Since each of these failed UA plans would correspond to a large number of TO failed plans, the solution density for TO would be considerably lower.

---

7. Since the depth-limit is equal to the length of the shortest solution, an iterative deepening (Korf, 1985) approach would yield similar results. Additionally, we note that increasing the depth-limit past the depth of the shortest solution does not significantly change the outcome of these experiments.

8. This definition of solution density is ill-defined for infinite trees, but we assume that a depth-bound is always provided, so only a finite subtree is explicitly enumerated.

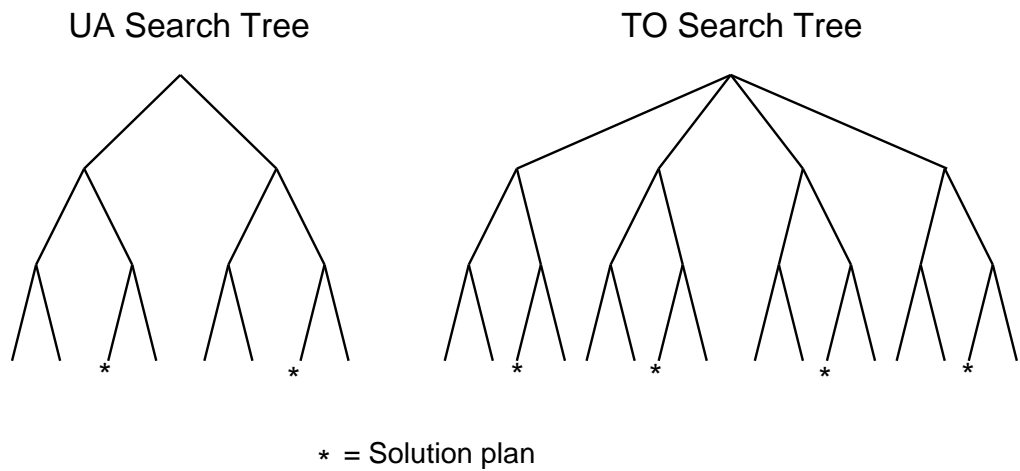

Figure 7: Uniform solution distribution, with solution density 0.25

For our blocksworld problems, we found that the solution densities of the two planners' trees does not differ greatly, at least not in such a way that would explain our performance results. We saw no tendency for $tree_{UA}$ to have a higher solution density than $tree_{TO}$. For example, for the 11 problems with solutions at depth six, the average solution density[9] for TO exceeded that of UA on 7 out of the 12 problems. This is not particularly surprising since we see no *a priori* reason to suppose that the solution densities of the two planners should differ greatly.

Since solution density is insufficient to explain UA's dominance on our blocksworld experiments when using depth-first search, we need to look elsewhere for an explanation. We hypothesize that the *distribution* of solutions provides an explanation. We note that if the solution plans are distributed perfectly uniformly (i.e., at even intervals) among the leaves of the search tree, and if the solution densities are similar, then both planners can be expected to search a similar number of leaves, as illustrated by the schematic search tree in Figure 7. Consequently, we can explain the observed dominance of UA over TO by hypothesizing that solutions are *not* uniformly distributed; that is, solutions tend to cluster. To see this, suppose that $tree_{UA}$ is smaller than $tree_{TO}$ but the two trees have the same solution density. If the solutions are clustered, as in Figure 8, then depth-first search can be expected to produce solutions more quickly for $tree_{UA}$ than for $tree_{TO}$.[10] The hypothesis that solutions tend to be clustered seems reasonable since it is easy to construct problems where a "wrong decision" near the top of the search tree can lead to an entire subtree that is devoid of solutions.

---

9. In our experiments, a nondeterministic goal selection procedure was used with our planners, which meant that the solution density could vary from run to run. We compared the average solution density over 25 trials for each problem to obtain our results.

10. Even if the solutions are distributed randomly amongst the leaves of the trees with uniform probability (as opposed to being distributed "perfectly uniformly"), there will be some clusters of nodes. Therefore, TO will have a small disadvantage. To see this, let us suppose that each leaf of both $tree_{UA}$ and $tree_{TO}$ is a solution with equal probability $p$. That is, if $tree_{UA}$ has $N_{UA}$ leaves, of which $k_{UA}$ are solutions,

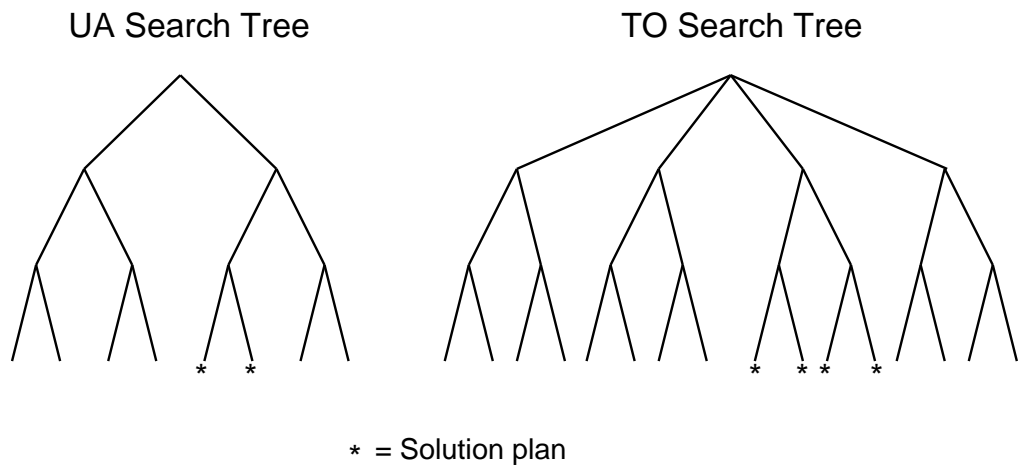

Figure 8: Non-uniform solution distribution, with solution density 0.25

One way to test our hypothesis is to compare UA and TO using a randomized search strategy, a type of Monte Carlo algorithm, that we refer to as "iterative sampling" (*cf.* Minton et al., 1992; Langley, 1992; Chen, 1989; Crawford & Baker, 1994). The iterative sampling strategy explores randomly chosen paths in the search tree until a solution is found. A path is selected by traversing the tree from the root to a leaf, choosing randomly at each branch point. If the leaf is a solution then search terminates; if not, the search process returns to the root and selects another path. The same path may be examined more than once since no memory is maintained between iterations.

In contrast to depth-first search, iterative sampling is relatively insensitive to the distribution of solutions. Therefore, the advantage of UA over TO should disappear if our hypothesis is correct. In our experiments, we did find that when UA and TO both use iterative sampling, they expand approximately the same number of nodes on our set of blocksworld problems.[11] (For both planners, performance with iterative sampling was worse than with depth-first search.) The fact that there is no difference between UA and TO under iterative sampling, but that there *is* a difference under depth-first search, suggests that solutions are

---

and $tree_{TO}$ has $N_{TO}$ leaves, of which $k_{TO}$ are solutions, then $p = k_{UA}/N_{UA} = k_{TO}/N_{TO}$. In general, if $k$ out of $N$ nodes are solutions, the expected number of nodes that must be tested to find a solution is $.5N/k$ when $k = 1$ and approaches $N/k$ as $k$ (and $N$) approaches $\infty$. (This is simply the expected number of samples for a binomial distribution.) Therefore, since $k_{TO} \geq k_{UA}$, the expected number of leaves explored by TO is greater than or equal to the expected number of leaves explored by UA, by at most a factor of 2.

11. The iterative sampling strategy was depth-limited in exactly the same way that our depth-first strategy was. We note, however, that the performance of iterative sampling is relatively insensitive to the actual depth-limit used.

indeed non-uniformly distributed. Furthermore, this result shows that UA is not necessarily superior to TO; the search strategy that is employed makes a dramatic difference.

Although our blocksworld domain may be atypical, we conjecture that our results are of general relevance. Specifically, for distribution-sensitive search strategies like depth-first search, one can expect that UA will tend to outperform TO. For distribution-insensitive strategies, such as iterative sampling, non-uniform distributions will have no effect. We note that while iterative sampling is a rather simplistic strategy, there are more sophisticated search strategies, such as iterative broadening (Ginsberg & Harvey, 1992), that are also relatively distribution insensitive. We further explore such strategies in Section 8.2.

## 8. The Role of Heuristics

In the preceding sections, we have shown that a partial-order planner can be more efficient simply because its search tree is smaller. With some search strategies, such as breadth-first search, this size differential obviously translates into an efficiency gain. With other strategies, such as depth-first search, the size differential translates into an efficiency gain, provided we make additional assumptions about the solution density and distribution.

However, it is often claimed that partial-order planners are more efficient due to their ability to make *more informed* ordering decisions, a rather different argument. For instance, Sacerdoti (1975) argues that this is the reason that NOAH performs well on problems such as the blocksworld's "Sussman anomaly". By delaying the decision of whether to stack A on B before or after stacking B on C, NOAH can eventually detect that a conflict will occur if it stacks A on B first, and a critic called "RESOLVE-CONFLICTS" can then order the steps intelligently.

In this section, we show that this argument can be formally described in terms of our two planners. We demonstrate that UA does in fact have a potential advantage over TO in that it can exploit certain types of heuristics more readily than TO. This advantage is independent of the fact that UA has a smaller search space. Whether or not this advantage is significant in practice is another question, of course. We also describe some experiments where we evaluate the effect of a commonly-used heuristic on our blocksworld problems.

### 8.1 Making More Informed Decisions

First, let us identify how it is that UA can make better use of certain heuristics than TO. In the UA planning algorithm, step 4 arbitrarily orders interacting plan steps. Similarly, Step 4 of TO arbitrarily chooses an insertion point for the new step. It is easy to see, however, that some orderings should be tried before others in a heuristic search. This is illustrated by Figure 9, which compares UA and TO on a particular problem. The key in the figure describes the relevant conditions of the library operators, where preconditions are indicated to the left of an operator and added conditions are indicated to the right (there are no deletes in this example). For brevity, the initial step and final step of the plans are not shown. Consider the plan in $tree_{UA}$ with unordered steps $O_1$ and $O_2$. When UA introduces $O_3$ to achieve precondition $p$ of $O_1$, Step 4 of UA will order $O_3$ with respect to $O_2$, since these steps interact. However, it makes more sense to order $O_2$ before $O_3$, since $O_2$ achieves precondition $q$ of $O_3$. This illustrates a simple planning heuristic that we refer to as the *min-goals* heuristic: "prefer the orderings that yield the fewest false preconditions".

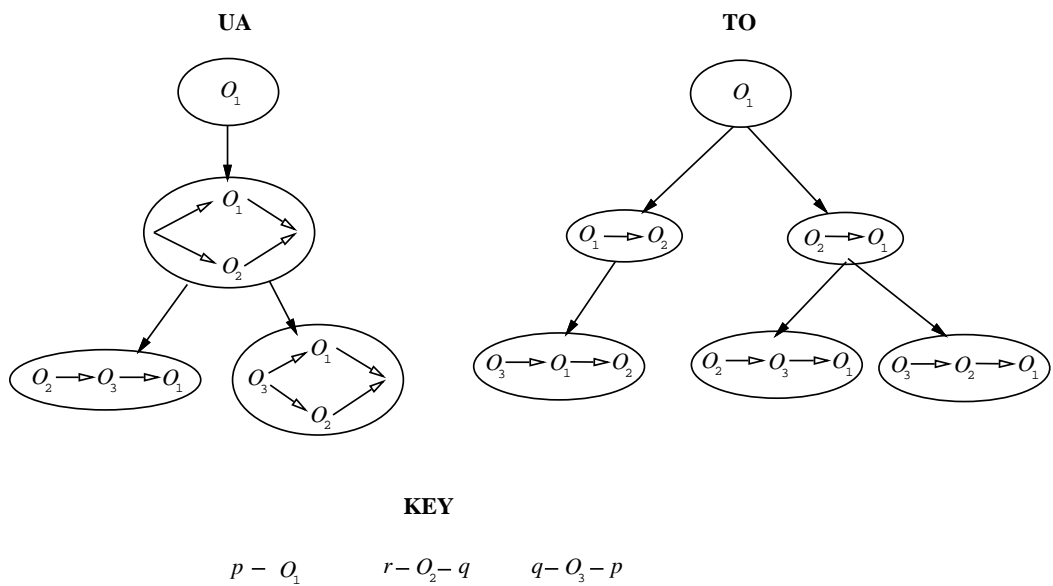

Figure 9: Comparison of UA and TO on an example.

This heuristic is not guaranteed to produce the optimal search or the optimal plan, but it is commonly used. It is the basis of the "resolve conflicts" critic that Sacerdoti employed in his blocksworld examples.

Notice, however, that TO cannot exploit this heuristic as effectively as UA because it prematurely orders $O_1$ with respect to $O_2$. Due to this inability to postpone an ordering decision, TO must choose arbitrarily between the plans $O_1 \prec O_2$ and $O_2 \prec O_1$, before the impact of this decision can be evaluated.

In the general case, suppose $h$ is a heuristic that can be applied to both partially ordered plans and totally ordered plans. Furthermore, assume $h$ is a "useful" heuristic; i.e., if $h$ rates one plan more highly than another, a planner that explores the more highly rated plan first will perform better on average. Then, UA will have a potential advantage over TO provided that $h$ satisfies the following property: for any UA plan $U$ and corresponding TO plan $T$, $h(U) \geq h(T)$; that is, a partially ordered plan must be rated at least as high as any of its linearizations. (Note that for unambiguous plans, the min-goals heuristic satisfies this property since it gives identical ratings to a partially ordered plan and its linearizations.)

UA has an advantage over TO because if UA is expanding plan $U$ and TO is expanding a corresponding plan $T$, then $h$ will rate some child of $U$ at least as high as the most highly rated child of $T$. This is true since every child of $T$ is a linearization of some child of $U$, and therefore no child of $T$ can be rated higher than a child of $U$. Furthermore, there may be a child of $U$ such that none of its linearizations is a child of $T$, and therefore this child of $U$ can be rated higher than every child of $T$. Since we assumed that $h$ is a useful heuristic, this means that UA is likely to make a better choice than TO.

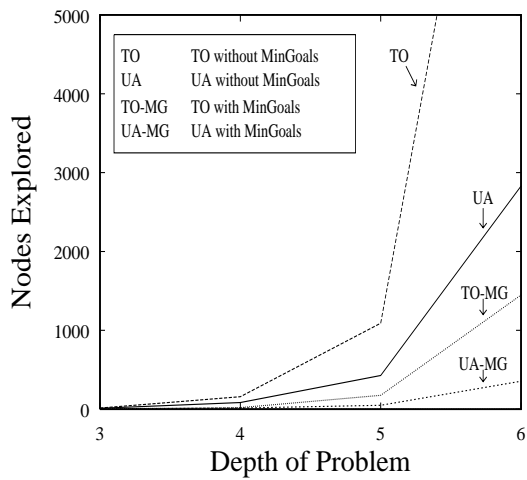

Figure 10: Depth first search with and without min-goals

## 8.2 Illustrative Experimental Results

The previous section showed that UA has a potential advantage over TO because it can better exploit certain ordering heuristics. We now examine the practical effects of incorporating one such heuristic into UA and TO.

First, we note that ordering heuristics only make sense for some search strategies. In particular, for breadth-first search, heuristics do *not* improve the efficiency of the search in a meaningful way (except possibly at the last level). Indeed, we need not consider any search strategy in which TO and UA are "synchronized", as defined earlier, since ordering heuristics do not significantly affect the relative performance of UA and TO under such strategies. Thus, we begin by considering a standard search strategy that is not synchronized: depth-first search.

We use the min-goals heuristic as the basis for our experimental investigation, since it is commonly employed, but presumably we could choose any heuristic that meets the criterion set forth in the previous section. Figure 10 shows the impact of min-goals on the behavior of UA and TO under depth-first search. Although the heuristic biases the order in which the two planners' search spaces are explored (*cf.* Rosenbloom, Lee, & Unruh, 1993), it appears that its effect is largely independent of the partial-order/total-order distinction, since both planners are improved by a similar percentage. For example, under depth-first search on the problems with solutions at depth six, UA improved 88% and TO improved 87%. Thus, there is no obvious evidence for any extra advantage for UA, as one might have expected from our analysis in the previous section. On the other hand, this does not contradict our theory, it simply means that the potential heuristic advantage was not significant enough to show up. In other domains, the advantage might manifest itself more significantly. After all, it is certainly possible to design problems in which the advantage is significant, as our example in Figure 9 illustrates. Our results simply illustrate that in our blocksworld domain, making intelligent ordering decisions produces a negligible advantage for UA, in contrast to the significant effect due to search space compression (discussed previously).[12]

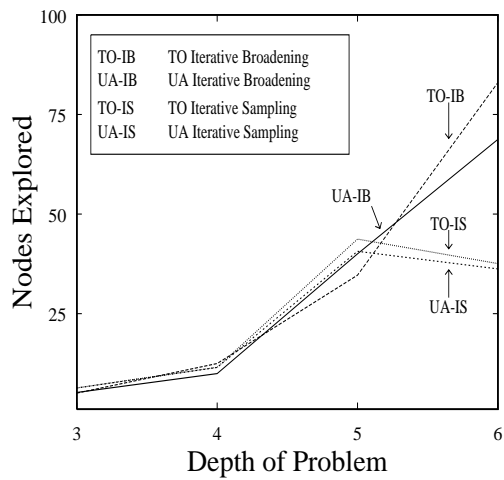

Figure 11: Iterative sampling & iterative broadening, both with min-goals

While the min-goals heuristic did not seem to help UA more than TO, the results are nevertheless interesting, since the heuristic had a very significant effect on the performance of *both* planners, so much so that TO *with* min-goals outperforms UA *without* min-goals. While the effectiveness of min-goals is domain dependent, we find it interesting that in these experiments, the use of min-goals makes more difference than the use of partial orders. After all, the blocksworld originally helped motivate the development of partial-order planning and most subsequent planning systems have employed partial orders. While not deeply surprising, this result does help reinforce what we already know: more attention should be paid to specific planning heuristics such as min-goals.

In our analysis of search space compression in Section 7, we described a "distribution insensitive" search strategy called iterative sampling and showed that under iterative sampling UA and TO perform similarly, although their performance is worse than it is under depth-first search. If we combine min-goals with iterative sampling, we find that this produces a much more powerful strategy, but one in which TO and UA still perform about equally. For simplicity, our implementation of iterative sampling uses min-goals as a pruning heuristic; at each choice point, it explores only those plan extensions with the fewest goals. This strategy is powerful, although incomplete.[13] Because of this incompleteness, we note there was one problem we removed from our sample set because iterative sampling with

---

12. In Section 9.2, we discuss planners that are "less-committed" than UA. For such planners, the advantage due to heuristics might be more pronounced since they "delay" their decisions even longer than UA.

13. Instead of exploring *only* those plan extensions with the fewest goals at each choice point, an alternative strategy is to assign each extension a probability that is inversely correlated with the number of goals, min-goals would never terminate on this problem. With this caveat in mind, we turn to the results in Figure 11, which when compared against Figure 10, show that the performance of both UA and TO with iterative sampling was, in general, significantly better than their performance under depth-first search. (Note that the graphs in Figures 10 and 11 have very different scales.) Our results clearly illustrate the utility of the planning bias introduced by min-goals in our blocksworld domain, since on 43 of our 44 problems, a solution exists in the very small subspace preferred by min-goals.

These experiments do not show any advantage for UA as compared with TO under the heuristic, which is consistent with our conclusions above. However, this could equally well be because min-goals was so powerful, leading to solutions so quickly, that smaller influences were obscured.

The dramatic success of combining min-goals with iterative sampling led us to consider another search strategy, iterative broadening, which combines the best aspects of depth-first search and iterative sampling. This more sophisticated search strategy initially behaves like iterative sampling, but evolves into depth-first search as the breadth-cutoff increases (Langley, 1992). Assuming that the solution is within the specified depth bound, iterative broadening is complete. In its early stages iterative broadening is distribution-insensitive; in its later stages it behaves like depth-first search and, thus, becomes increasingly sensitive to solution distribution. As one would expect from our iterative sampling experiments, with iterative broadening, solutions were found very early on, as shown in Figure 11. Thus, it is not surprising that UA and TO performed similarly under iterative broadening.

We should point out that the results presented in this subsection are only illustrative, since they deal with only a single domain and with a single heuristic. Nevertheless, our experiments do illustrate how the various properties we have identified in this paper can interact.

## 9. Extending our Results

Having established our basic results concerning the efficiency of UA and TO under various circumstances, we now consider how these results extend to other types of planners.

### 9.1 More Expressive Languages

In the preceding sections, we showed that the primary advantage that UA has over TO is that UA's search tree may be exponentially smaller than TO's search tree, and we also showed that UA only pays a small (polynomial) extra cost per node for this advantage. Thus far we have assumed a very restricted planning language in which the operators are propositional; however, most practical problems demand operators with variables, conditional effects, or conditional preconditions. With a more expressive planning language, will the time cost per node be significantly greater for UA than for TO? One might think so, since the work required to identify interacting steps can increase with the expressiveness of the operator language used (Dean & Boddy, 1988; Hertzberg & Horz, 1989). If the cost of detecting step

---

and pick accordingly. Given a depth bound, this strategy has the advantage of being *asymptotically complete*. We used the simpler strategy here for pedagogical reasons.

interaction is high enough, the savings that UA enjoys due to its reduced search space will be outweighed by the additional expense incurred at each node.

Consider the case for simple breadth-first search. Earlier we showed that the ratio of the total time costs of TO and UA is as follows, where the subtree considered by UA under breadth-first search is denoted by $bf(tree_{UA})$, the number of steps in plan a $U$ is denoted by $n_u$, and the number of edges in $U$ is denoted by $e_u$:

$$\frac{\text{cost}(\text{TO}_{bf})}{\text{cost}(\text{UA}_{bf})} = \frac{\sum_{U \in bf(tree_{UA})} O(n_u) \cdot \mid \mathcal{L}(U) \mid}{\sum_{U \in bf(tree_{UA})} O(e_u)}$$

This cost comparison is specific to the simple propositional operator language used so far, but the basic idea is more general. UA will generally outperform TO whenever its cost per node is less than the product of the cost per node for TO and the number of TO nodes that correspond under $\mathcal{L}$. Thus, UA could incur an exponential cost per node and still outperform TO in some cases. This can happen, for example, if the exponential number of linearizations of a UA partial order is greater than the exponential cost per node for UA. In general, however, we would like to avoid the case where UA pays an exponential cost per node and, instead, consider an approach that can guarantee that the cost per node for UA remains polynomial (as long as the cost per node for TO also remains polynomial).

The cost per node for UA is dominated by the cost of updating the goal set (Step 5) and the cost of selecting the orderings (Step 4). Updating the goal set remains polynomial as long as a plan is unambiguous. Since each precondition in an unambiguous plan is either *necessarily* true or *necessarily* false, we can determine the truth value of a given precondition by examining its truth value in an arbitrary linearization of the plan. Thus, we can simply linearize the plan and then use the same procedure TO uses for calculating the goal set. As a result, it is only the cost of maintaining the unambiguous property (i.e., Step 4) that is impacted by more expressive languages. One approach for efficiently maintaining this property relies on a "conservative" ordering strategy in which operators are ordered if they even *possibly* interact.

As an illustration of this approach, consider a simple propositional language with conditional effects, such as "if $p$ and $q$, then add $r$". Hence, an operator can add (or delete) propositions depending on the state in which it is applied. We refer to conditions such as "$p$" in our example as *dependency conditions*. (Note that, like preconditions, dependency conditions are simple propositions.) Chapman (1987) showed that with this type of language it is NP-hard to decide whether a precondition is true in a partially ordered plan. However, as we pointed out above, for the special case of *unambiguous* plans, this decision can be accomplished in polynomial time.

Formally, the language is specified as follows. An operator $O$, as before, has a list of preconditions, pre($O$), a list of (unconditional) adds, adds($O$), a list of (unconditional) deletes, dels($O$). In addition, it has a list of conditional adds, cadds($O$), and a list of conditional deletes, cdels($O$); both containing pairs $\langle D_e, e \rangle$, where $D_e$ is a conjunctive set of dependency conditions and $e$ is the conditional effect (either an added or a deleted condition). Analogous with the constraint that every delete must be a precondition, every conditional delete must be a member of its dependency conditions; that is, for every $\langle D_e, e \rangle \in$ cdels($O$), $e \in D_e$.

Figure 12 shows a version of the UA algorithm, called UA-C, which is appropriate for this language. The primary difference between the UA and UA-C algorithms is that in both Steps 3 and 4b an operator may be *specialized* with respect to a set of dependency conditions. The function specialize($O$, $D$) accepts a plan step, $O$, and a set of dependency conditions, $D$; it returns a new step $O'$ that is just like $O$, but with certain conditional effects made unconditional. The effects that are selected for this transformation are exactly those whose dependency conditions are a subset of $D$. Thus, the act of specializing a plan step is the act of committing to expanding its causal role in a plan.[14] Once a step is specialized, UA-C has made a commitment to use it for a given set of effects. Of course, a step can be further specialized in a later search node, but specializations are never retracted.

More precisely, the definition of $O' = \text{specialize}(O, D)$, where $O$ is a step, $D$ is a conjunctive set of dependency conditions in $O$, and $\setminus$ is the set difference operator, is as follows.

- $\text{pre}(O') = \text{pre}(O) \cup D$.

- $\text{adds}(O') = \text{adds}(O) \cup \{e \mid \langle D_e, e \rangle \in \text{cadds}(O) \wedge D_e \subset D\}$.

- $\text{dels}(O') = \text{dels}(O) \cup \{e \mid \langle D_e, e \rangle \in \text{cdels}(O) \wedge D_e \subset D\}$.

- $\text{cadds}(O') = \{\langle D_{e'}, e \rangle \mid \langle D_e, e \rangle \in \text{cadds}(O) \wedge D_e \not\subset D \wedge D_{e'} = D_e \setminus D\}$.

- $\text{cdels}(O') = \{\langle D_{e'}, e \rangle \mid \langle D_e, e \rangle \in \text{cdels}(O) \wedge D_e \not\subset D \wedge D_{e'} = D_e \setminus D\}$.

The definition of step interaction is generalized for UA-C as follows. We say that two steps in a plan *interact* if they are unordered with respect to each other and the following disjunction holds:

- one step has a precondition *or dependency condition* that is added or deleted by the other step, or

- one step adds a condition that is deleted by the other step.

The difference between this definition of step interaction and the one given earlier is indicated by an italic font. This modified definition allows us to detect interacting operators with a simple inexpensive test, as did our original definition. For example, two steps that are unordered interact if one step conditionally adds $r$ and the other has precondition $r$. Note that the first step need not actually add $r$ in the plan, so ordering the two operators might be unnecessary. In general, our definition of interaction is a sufficient criterion for guaranteeing that the resulting plans are unambiguous, but it is not a necessary criterion.

Figure 13 shows a schematic example illustrating how UA-C extends a plan. The preconditions of each operator are shown on the left of each operator, and the unconditional adds on the right. (We only show the preconditions and effects necessary to illustrate the specialization process; no deletes are used in the example.) Conditional adds are shown

---

14. For simplicity, the modifications used to create UA-C are not very sophisticated. As a result, UA-C's space may be larger than it needs to be in some circumstances, since it aggressively commits to specializations. A more sophisticated set of modifications is possible; however, the subtlies involved in efficiently planning with dependency conditions (Pednault, 1988; Collins & Pryor, 1992; Penberthy & Weld, 1992) are largely irrelevant to our discussion.

UA-C$(P, G)$

1. **Termination check:** If $G$ is empty, report success and return solution plan P.

2. **Goal selection:** Let $c =$ select-goal$(G)$, and let $O_{need}$ be the plan step for which $c$ is a precondition.

3. **Operator selection:** Let $O_{add}$ be an operator schema in the library that *possibly* adds $c$; that is, either $c \in$ adds$(O)$, or there exists an $\langle D_c, c \rangle \in$ cadds$(O)$. In the former case, insert step $O_{add}$ and in the latter case, insert step specialize$(O_{add}, D_c)$. If there is no such $O_{add}$, then terminate and report failure. *Choice point: all ways in which c can be added must be considered for completeness.*

4a. **Ordering selection:** Let $O_{del}$ be the (unconditional) last deleter of $c$. Order $O_{add}$ after $O_{del}$ and before $O_{need}$.

   Repeat until there are no interactions:
   - Select a step $O_{int}$ that interacts with $O_{add}$.
   - Order $O_{int}$ either before or after $O_{add}$.
     *Choice point: both orderings must be considered for completeness.*

   Let $P'$ be the resulting plan.

4b. **Operator role selection:** While there exists a step $O_{cadd}$ with unmarked conditional add $\langle D_c, c \rangle$ and a step $O_{use}$ with precondition $c$, such that $O_{use}$ is after $O_{cadd}$ and there is no (unconditional) deleter of $c$ in between $O_{use}$ and $O_{cadd}$.
   - Either mark $\langle D_c, c \rangle$, or replace $O_{cadd}$ with specialize$(O_{cadd}, D_c)$.
     *Choice point: Both options must be considered for completeness.*

5. **Goal updating:** Let $G'$ be the set of preconditions in $P'$ that are necessarily false.

6. **Recursive invocation:** UA-C$(P', G')$.

Figure 12: The UA-C planning algorithm

underneath each operator. For instance, the first operator in the plan at the top of the page has precondition $p$. This operator adds $q$ and conditionally adds $u$ if $t$ is true. The figure illustrates two of the plans produced as a result of adding a new conditional operator to the plan. In one plan, the conditional effects $[u \to s]$ and $[t \to u]$ are selected in the specialization process, and in the other plan they are not.

The new step, Step 4b, requires only polynomial time per plan generated, and the time cost of the other steps are the same as for UA. Hence, as with our original UA algorithm, the cost per node for the UA-C algorithm is polynomial.

TO can also handle this language given the corresponding modifications (changing Step 3 and adding Step 4b), and the time cost per plan also remains polynomial.[15] Moreover, the same relationship holds between the two planners' search spaces − $tree_{UA}$ is never larger than $tree_{TO}$ and can be exponentially smaller. This example illustrates that the theoretical advantages that UA has over TO can be preserved for a more expressive language. As we pointed out, our definition of interaction is a sufficient criterion for guaranteeing that the resulting plans are unambiguous, but it is not a necessary criterion. Nevertheless, this conservative approach allows interactions to be detected via a simple inexpensive syntactic test. Essentially, we have kept the cost per node for UA-C low by restricting the search space it considers, as shown in Figure 14. UA-C only considers unambiguous plans that can be generated via its "conservative" ordering strategy. UA-C is still a partial-order planner, and it is complete, but it does not consider all partially ordered plans or even all unambiguous partially ordered plans.

---

15. In fact, Step 4b be implemented so that the time cost is $O(e)$, using the graph traversal techniques described in Section 6. As a result the UA-C implementation and the corresponding TO-C implementation have the same time cost per node for this new language as they did for the original language, $O(e)$ and $O(n)$, respectively.

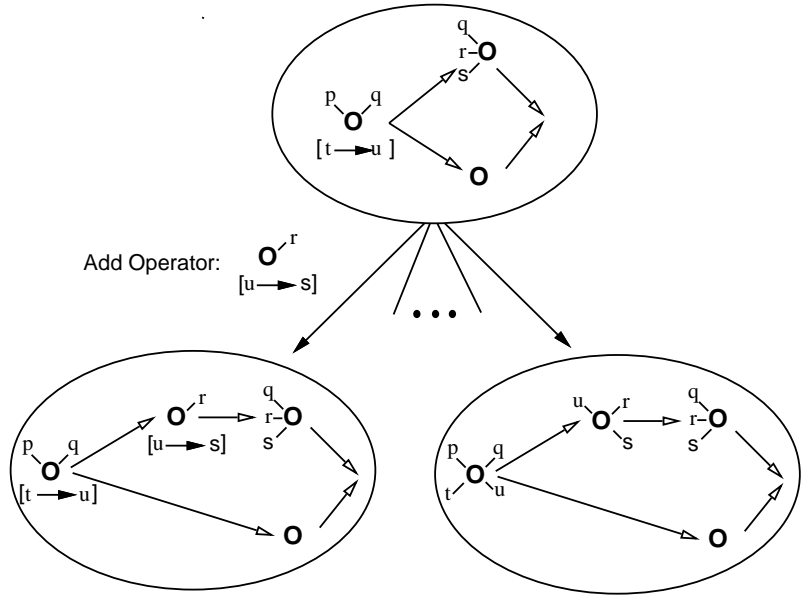

Figure 13: An example illustrating the UA-C algorithm

The same "trick" can be used for other languages as well, provided that we can devise a simple test to detect interacting operators. For example, in previous work (Minton et al., 1991b) we showed how this can be done for a language where operators can have variables in their preconditions and effects. In the general case, for a given UA plan and a corresponding TO plan, Steps 1,2, and 3 of the UA algorithm cost the same as the corresponding steps of the TO algorithm. As long as the plans considered by UA are unambiguous, Step 5 of the UA algorithm can be accomplished with an arbitrary linearization of the plan, in which case it costs at most $O(e)$ more than Step 5 of the TO algorithm. Thus, the only possibility for additional cost is in Step 4. In general, if we can devise a "local" criterion for interaction such that the resulting plan is guaranteed to be unambiguous, then the ordering selection step can be accomplished in polynomial time. By "local", we mean a criterion that only considers operator pairs to determine interactions; i.e., it must not examine the rest of the plan.

Although the theoretical advantages that UA has over TO can be preserved for more expressive languages, there is a cost. The unambiguous plans that are considered may have more orderings than necessary, and the addition of unnecessary orderings can increase the size of UA's search tree. The magnitude of this increase depends on the specific language, domain, and problem being considered. Nevertheless, we can guarantee that UA's search tree is never larger than TO's.

The general lesson here is that the cost of plan extension is not solely dependent on the expressiveness of the operator language, it also depends on the nature of the plans that the planner considers. So, although the extension of partially ordered plans is NP-hard for languages with conditional effects, if the space of plans is restricted (e.g., only unambiguous plans are considered) then this worst-case situation is avoided.

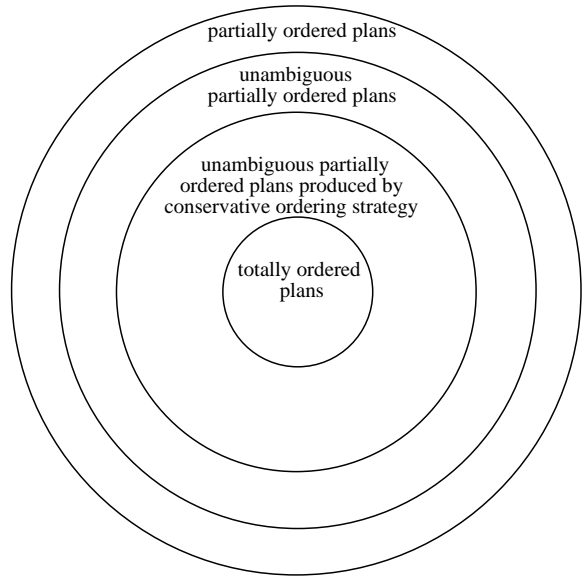

Figure 14: Hierarchy of Plan Spaces

## 9.2 Less Committed Planners

We have shown that UA, a partial-order planner, can have certain computational advantages over a total-order planner, TO, since its ability to delay commitments allows for a more compact search space and potentially more intelligent ordering choices. However, there are many planners that are even less committed than UA. In fact, there is a continuum of commitment strategies that we might consider, as illustrated in Figure 15. Total-order planning lies at one end of the spectrum. At the other extreme is the strategy of maintaining a *totally unordered* set of steps during search until there exists a linearization of the steps that is a solution plan.

Compared to many well-known planners, UA is conservative since it requires each plan to be unambiguous. This is not required by NOAH (Sacerdoti, 1977), NonLin (Tate, 1977),

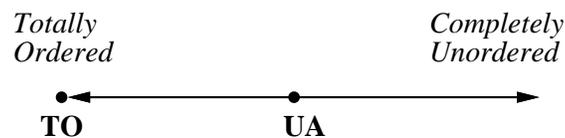

Figure 15: A continuum of commitment strategies

MT($P$, $G$)

1. **Termination check:** If $G$ is empty, report success and stop.

2. **Goal selection:** Let $c$ = select-goal($G$), and let $O_{need}$ be the plan step for which $c$ is a precondition.

3. **Operator selection:** Let $O_{add}$ be either a plan step possibly before $O_{need}$ that adds $c$ or an operator in the library that adds $c$. If there is no such $O_{add}$, then terminate and report failure.
   *Choice point: all such operators must be considered for completeness.*

4. **Ordering selection:** Order $O_{add}$ before $O_{need}$. Repeat until there are no steps possibly between $O_{add}$ and $O_{need}$ which delete $c$:
   Let $O_{del}$ be such a step; choose one of the following ways to make $c$ true for $O_{need}$
   ○ Order $O_{del}$ after $O_{need}$.
   ○ Choose a step $O_{knight}$ (possibly $O_{add}$) that adds $c$ that is possibly between $O_{del}$ and $O_{need}$; order it after $O_{del}$ and before $O_{need}$.
   *Choice point: both alternatives must be considered for completeness.*
   Let $P'$ be the resulting plan.

5. **Goal updating:** Let $G'$ be the set of preconditions in $P'$ that are not necessarily true.

6. **Recursive invocation:** MT($P'$, $G'$).

Figure 16: A Propositional Planner based on the Modal Truth Criterion

nor Tweak (Chapman, 1987), for example. How do these less-committed planners compare to UA and TO? One might expect a less-committed planner to have the same advantages over UA that UA has over TO. However, this is not necessarily true. As an example, in this section we introduce a Tweak-like planner, called MT, and show that its search space is larger than even TO's in some circumstances.[16]

Figure 16 presents the MT procedure. MT is a propositional planner based on Chapman's Modal Truth Criterion (Chapman, 1987), the formal statement that characterizes Tweak's search space. It is straightforward to see that MT is less committed than UA. The algorithms are quite similar; however, in Step 4, whereas UA orders all interacting steps, MT does not. Since MT does not immediately order all interacting operators, it may have to add additional orderings between previously introduced operators later in the planning process to produce correct plans.

The proof that UA's search tree is no larger than TO's search tree rested on the two properties of $\mathcal{L}$ elaborated in Section 5. By investigating the relationship between MT and TO, we found that the second property, the disjointness property, does not hold for MT, and its failure illustrates how MT can explore more plans than TO (and, consequently, than UA) on certain problems. The disjointness property guarantees that UA does not generate "overlapping" plans. The example in Figure 17 shows that MT fails to satisfy this property because it can generate plans that share common linearizations, leading to considerable redundancy in the search tree. The figure shows three steps, $O_1$, $O_2$, and $O_3$, where each $O_i$ has precondition $p_i$ and added conditions $g_i$, $p_1$, $p_2$, and $p_3$. The final step has preconditions $g_1$, $g_2$, and $g_3$, but the initial and final steps are not shown in the figure. At the top of the figure, in the plan constructed by MT, goals $g_1$, $g_2$, and $g_3$ have been achieved, but $p_1$, $p_2$, and $p_3$ remain to be achieved. Subsequently, in solving precondition $p_1$, MT generates plans which share the linearization $O_3 \prec O_2 \prec O_1$ (among others). In comparison, both TO and UA only generate the plan $O_3 \prec O_2 \prec O_1$ once. In fact, it is simple to show that, under breadth-first search, MT explores many more plans than TO on this example (and also more than UA, by transitivity) due to the redundancy in its search space.

---

16. We use Tweak for this comparison because, like UA and TO, it is a formal construct rather than a realistic planner, and therefore more easily analyzed.

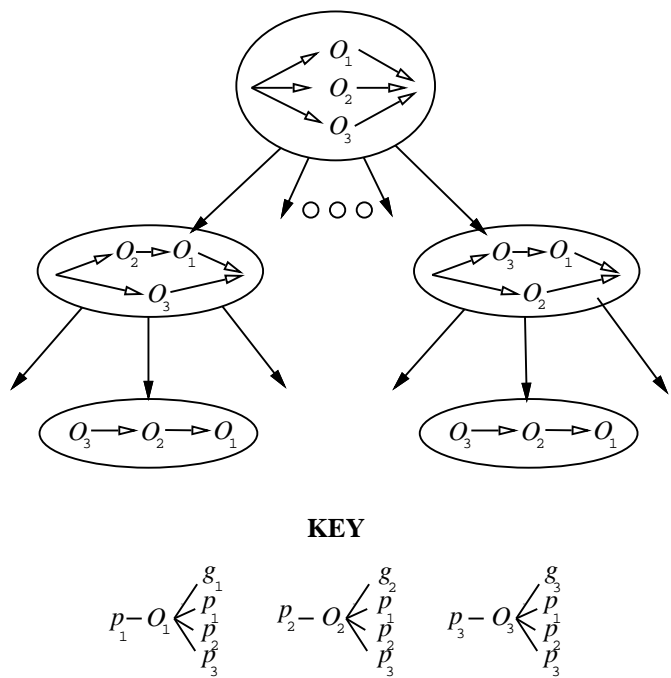

Figure 17: "Overlapping" plans.

This result may seem counterintuitive. However, note that the search space size for a partial-order planner is potentially much greater than that of a total-order planner since there are many more partial orders over a set of steps than there are total orders. (Thus, when designing a partial-order planner, one may preclude overlapping linearizations in order to avoid redundancy, as discussed by McAllester & Rosenblitt, 1991 and Kambhampati, 1994c.)

Of course, one can also construct examples where MT does have a smaller search space than both UA and TO. Our example simply illustrates that although one planner may be less committed than another, its search space is not *necessarily smaller*. The commitment strategy used by a planner is simply one factor that influences overall performance. In particular, the effect of redundancy in a partial-order planner can overwhelm other considerations. In comparing two planners, one must carefully consider the mapping between their search spaces before concluding that "less committed $\Rightarrow$ smaller search space".

## 10. Related Work

For many years, the intuitions underlying partial-order planning were largely taken for granted. Only in the past few years has there been renewed interest in the fundamental principles underlying these issues.

Barrett et al. (1991) and Barrett and Weld (1994) describe an interesting and novel analysis of partial-order planning that complements our own work. They compare a partial-order planner with two total-order planners derived from it, one that searches in the space of plans, and the other that searches in the space of world states. Their study focuses on how the goal structure of the problem affects the efficiency of partial-order planning. Specifically, they examine how partial-order and total-order planning compare for problems with independent, serializable, and non-serializable goals, when using a resource-bounded depth-first search. They refine Korf's work on serializable goals (Korf, 1987), introducing a distinction between trivially serializable subgoals, where the subgoals can be solved in any order without violating a previously solved subgoal, and laboriously serializable subgoals, where the subgoals are serializable, but at least $1/n$ of the orderings can cause a previously solved subgoal to be violated. Their study describes conditions under which a partial-order planner may have an advantage. For instance, they show that in a domain where the goals are trivially serializable for their partial-order planner and laboriously serializable for their total-order planners, their partial-order planner performs significantly better.

Our study provides an interesting contrast to Barret and Weld's work, since we investigate the relative efficiencies of partial-order and total-order planning algorithms independent of any particular domain structure. Instead, we focus on the underlying properties of the search space and how the search strategy affects the efficiency of our planners. Nevertheless, we believe there are interesting relationships between the forms of serializability that they investigate, and the ideas of solution density and clustering that we have discussed here. To illustrate this, consider an artificial domain that Barret and Weld refer to as $D^1S^1$, where, in each problem, the goals are a subset of $\{G_1, G_2, \ldots G_{15}\}$, the initial conditions are $\{I_1, I_2, \ldots I_{15}\}$, and each operator $O_{i \in \{1,2,\ldots,15\}}$ has precondition $I_i$, adds $G_i$, and deletes $I_{i-1}$. It follows that if a solution in $D^1S^1$ contains operators $O_i$ and $O_j$ where $i < j$, then $O_i$ must precede $O_j$. In this domain, the goals are trivially serializable for their partial-order planner and laboriously serializable for their total-order planners; thus, the partial-order planner performs best. But note also that in this artificial domain, there is exactly one solution per problem and it is totally ordered. Therefore, it is immediately clear that, if we give UA and TO problems from this domain, then UA's search tree will generally be much smaller than TO's search tree. Since there is only single solution for both planners, the solution density for UA will clearly be greater than that for TO. Thus, the properties we discussed in this paper should provide a basis for analyzing how differences in subgoal serializibility manifest their effect on the search. This subject, however, is not as simple as it might seem and deserves further study.

In other related work, Kambhampati has written several papers (Kambhampati, 1994a, 1994b, 1994c) that analyze the design space of partial-order planners, including the UA planner presented here. Kambhampati compares UA, Tweak, SNLP (McAllester & Rosenblitt, 1991), UCPOP (Penberthy & Weld, 1992), and several other planners along a variety of dimensions. He presents a generalized schema for partial order planning algorithms (Kambhampati, 1994c) and shows that the commitment strategy used in UA can be viewed as a way to increase the tractability of the plan extension (or refinement) process. Kambhampati also carries out an empirical comparison of the various planning algorithms on a particular problem (Kambhampati, 1994a), showing how the differences in commitment strategies affects the efficiency of the planning process. He distinguishes two separate components of the branching factor, $b_t$ and $b_e$, the former resulting from the commitment strategy for operator ordering (or in his terms, the "tractability refinements") and the latter resulting from the choice of operator ("establishment refinements"). Kambhampati's experiments demonstrate that while "eager" commitment strategies tend to increase $b_t$, sometimes they also decrease $b_e$, because the number of possible establishers is reduced when plans are more ordered. This is, of course, closely related to the issues investigated in this paper.

In addition, Kambhampati and Chen (1993) have compared the relative utility of *reusing* partially ordered and totally ordered plans in "learning planners". They showed that the reuse of partially ordered plans, rather than totally ordered plans, result in "storage compaction" because they can represent a large number of different orderings. Moreover, partial-order planners have an advantage because they can exploit such plans more effectively than total-order planners. In many respects, these advantages are fundamentally similar to the advantages that UA derives from its potentially smaller search space.

## 11. Conclusions

By focusing our analysis on a single issue, namely, operator ordering commitment, we have been able to carry out a rigorous comparative analysis of two planners. We have shown that the search space of a partial-order planner, UA, is *never* larger than the search space of a total-order planner, TO. Indeed for certain problems, UA's search space is exponentially smaller than TO's. Since UA pays only a small polynomial time increment per node over TO, it is generally more efficient.

We then showed that UA's search space advantage may not necessarily translate into an efficiency gain, depending in subtle ways on the search strategy and heuristics that are employed by the planner. For example, our experiments suggest that distribution-sensitive search strategies, such as depth-first search, can benefit more from partial orders than can search strategies that are distribution-insensitive.

We also examined a variety of extensions to our planners, in order to demonstrate the generality of these results. We argued that the potential benefits of partial-order planning may be retained even with highly expressive planning languages. However, we showed that partial-order planners do not necessarily have smaller search spaces, since some "less-committed" strategies may create redundancies in the search space. In particular, we demonstrated that a Tweak-like planner, MT, can have a larger search space than our total-order planner on some problems.

How general are these results? Although our analysis has considered only two specific planners, we have examined some important tradeoffs that are of general relevance. The analysis clearly illustrates how the planning language, the search strategy, and the heuristics that are used can affect the relative advantages of the two planning styles.

The results in this paper should be considered as an investigation of the possible benefits of partial-order planning. UA and TO have been constructed in order for us to analyze the total-order/partial-order distinction in isolation. In reality, the comparative behavior of two planners is rarely as clear (as witnessed by our discussion of MT). While the general points we make are applicable to other planners, if we chose two arbitrary planners, we would not expect one planner to so clearly dominate the other.

Our observations regarding the interplay between plan representation and search strategy raise new concerns for comparative analyses of planners. Historically, it has been assumed that representing plans as partial orders is categorically "better" than representing plans as total orders. The results presented in this paper begin to tell a more accurate story, one that is both more interesting and more complex than we initially expected.

## Appendix A. Proofs

### A.1 Definitions

This section defines the terminology and notation used in our proofs. The notion of plan equivalence is introduced here because each plan step is, by definition, a uniquely labeled operator instance, as noted in Section 3 and Section 5. Thus, no two plans have the same set of steps. Although this formalism simplifies our analysis, it requires us to define plan equivalence explicitly.

- A *plan* is a pair $\langle \theta, \prec \rangle$, where $\theta$ is a set of steps, and $\prec$ is the "before" relation on $\theta$, i.e., $\prec$ is a *strict partial order* on $\theta$. Notationally, $O_1 \prec O_2$ if and only if $(O_1, O_2) \in \prec$.

- For a given problem, we define the search tree $tree_{TO}$ as the complete tree of plans that is generated by the TO algorithm on that problem. $tree_{UA}$ is the corresponding search tree generated by UA on the same problem.

- Two plans, $P_1 = \langle \theta_1, \prec_1 \rangle$ and $P_2 = \langle \theta_2, \prec_2 \rangle$ are said to be *equivalent*, denoted $P_1 \simeq P_2$, if there exists a bijective function $f$ from $\theta_1$ to $\theta_2$ such that:

    - for all $O \in \theta_1$, $O$ and $f(O)$ are instances of the same operator, and
    - for all $O', O'' \in \theta_1$, $O' \prec O''$ if and only if $f(O') \prec f(O'')$.

- A plan $P_2$ is a *1-step TO-extension* (or *1-step UA-extension*) of a plan $P_1$ if $P_2$ is equivalent to some plan produced from $P_1$ in one invocation of TO (or UA).

- A plan $P$ is a *TO-extension* (or *UA-extension*) if either:

    - $P$ is the initial plan, or
    - $P$ is a 1-step TO-extension (or 1-step UA-extension) of a TO-extension (or UA-extension).

    It immediately follows from this definition that if $P$ is a member of $tree_{TO}$ (or $tree_{UA}$), then $P$ is a TO-extension (or UA-extension). In addition, if $P$ is a TO-extension (or UA-extension), then some plan that is equivalent to $P$ is a member of $tree_{TO}$ (or $tree_{UA}$).

- $P_1$ is a *linearization of* $P_2 = \langle \theta, \prec_2 \rangle$ if there exists a strict total order $\prec_1$ such that $\prec_2 \subseteq \prec_1$ and $P_1 \simeq \langle \theta, \prec_1 \rangle$.

- Given a search tree, let *parent* be a function from a plan to its parent plan in the tree. Note that $P_1$ is the parent of $P_2$, denoted $P_1 = parent(P_2)$, only if $P_2$ is a 1-step extension of $P_1$.

- Given $U \in tree_{UA}$ and $T \in tree_{TO}$, $T \in \mathcal{L}(U)$ if and only if plan $T$ is a linearization of plan $U$ and either both $U$ and $T$ are root nodes of their respective search trees, or $parent(T) \in \mathcal{L}(parent(U))$.

- The *length* of the plan is the number of steps in the plan excluding the first and last steps. Thus, the initial plan has length 0. A plan $P$ with $n$ steps has length $n - 2$.

- $P_1$ is a *subplan of* $P_2 = \langle \theta_2, \prec_2 \rangle$ if $P_1 \simeq \langle \theta_1, \prec_1 \rangle$, where

  - $\theta_1 \subseteq \theta_2$ and
  - $\prec_1$ is $\prec_2$ restricted to $\theta_1$, i.e., $\prec_1 = \prec_2 \cap \theta_1 \times \theta_1$.

- $P_1$ is a *strict* subplan of $P_2$, if $P_1$ is a subplan of $P_2$ and the length of $P_1$ is less than the length of $P_2$.

- A solution plan $P$ is a *compact solution* if no strict subplan of $P$ is a solution.

## A.2 Extension Lemmas

**TO-Extension Lemma:** Consider totally ordered plans $T_0 = \langle \theta_0, \prec_0 \rangle$ and $T_1 = \langle \theta_1, \prec_1 \rangle$, such that $\theta_1 = \theta_0 \cup \{O_{add}\}$ and $\prec_0 \subset \prec_1$. Let $G$ be the set of false preconditions in $T_0$. Then $T_1$ is a 1-step TO-extension of $T_0$ if:

- $c = \text{select-goal}(G)$, where $c$ is the precondition of some step $O_{need}$ in $T_0$, and

- $O_{add}$ adds $c$, and

- $(O_{add}, O_{need}) \in \prec_1$, and

- $(O_{del}, O_{add}) \in \prec_1$, where $O_{del}$ is the last deleter of $c$ in $T_1$.

**Proof Sketch:** This lemma follows from the definition of TO. Given plan $T_0$, with false precondition $c$, once TO selects $c$ as the goal, TO will consider all operators that achieve $c$, and for each operator TO considers all positions before $c$ and after the last deleter of $c$.

**UA-Extension Lemma:** Consider a plan $U_0 = \langle \theta_0, \prec_0 \rangle$ produced by UA and plan $U_1 = \langle \theta_1, \prec_1 \rangle$, such that $\theta_1 = \theta_0 \cup \{O_{add}\}$ and $\prec_0 \subset \prec_1$. Let $G$ be the set of false preconditions of the steps in $U_0$. Then $U_1$ is a 1-step UA-extension of $U_0$ if:

- $c = \text{select-goal}(G)$, where $c$ is the precondition of some step $O_{need}$ in $U_0$, and

- $O_{add}$ adds $c$, and

- $\prec_1$ is a minimal set of consistent orderings such that

  - $\prec_0 \subseteq \prec_1$, and
  - $(O_{add}, O_{need}) \in \prec_1$, and
  - $(O_{del}, O_{add}) \in \prec_1$, where $O_{del}$ is the last deleter of $c$ in $U_1$, and
  - no step in $U_1$ interacts with $O_{add}$

**Proof Sketch:** This lemma follows from the definition of UA. Given plan $U_0$, with false precondition $c$, UA considers all operators that achieve $c$, and for each such operator UA then inserts it in the plan such that it is before $c$ and after the last deleter. UA then considers all consistent combinations of orderings between the new operator and the operators with which it interacts. No other orderings are added to the plan.

## A.3 Proof of Search Space Correspondence $\mathcal{L}$

**Mapping Lemma:** Let $U_0 = \langle \theta_0, \prec_{u0} \rangle$ be an unambiguous plan and let $U_1 = \langle \theta_1, \prec_{u1} \rangle$ be a 1-step UA-extension of $U_0$. If $T_1 = \langle \theta_1, \prec_{t1} \rangle$ is a linearization of $U_1$, then there exists a plan $T_0$ such that $T_0$ is a linearization of $U_0$ and $T_1$ is a 1-step TO-extension of $T_0$.

**Proof:** Since $U_1$ is a 1-step UA-extension of $U_0$, there is a step $O_{add}$ such that $\theta_1 = \theta_0 \cup \{O_{add}\}$. Let $T_0$ be the subplan produced by removing $O_{add}$ from $T_1$; that is, $T_0 = \langle \theta_0, \prec_{t0} \rangle$, where $\prec_{t0} = \prec_{t1} \cap \theta_0 \times \theta_0$. Since $\prec_{u0} = \prec_{u1} \cap \theta_0 \times \theta_0 \subseteq \prec_{t1} \cap \theta_0 \times \theta_0 = \prec_{t0}$, it follows that $T_0$ is a linearization of $U_0$.

Using the TO-Extension lemma, we can show that $T_1$ is a 1-step TO-extension of $T_0$. First, $T_0$ is a linearization of $U_0$, so the two plans have the same set of goals. Therefore, if UA selects some goal $c$ in expanding $U_0$, TO selects $c$ in extending $T_0$. Second, it must be the case that $O_{add}$ adds $c$ since $O_{add}$ is the step UA inserted into $U_0$ to make $c$ true. Third, $O_{add}$ is before $O_{need}$ in $T_1$, since $O_{add}$ is before $O_{need}$ in $U_1$ (by definition of UA) and since $T_1$ is a linearization of $U_1$. Fourth, $O_{add}$ is after the last deleter of $c$, $O_{del}$, in $T_1$, since $O_{add}$ is after $O_{del}$ in $U_1$ (by definition of UA) and since $T_1$ is a linearization of $U_1$. Therefore, the conditions of the TO-Extension lemma hold and, thus, $T_1$ is a 1-step TO-extension of $T_0$. *Q.E.D.*

**Totality Property** For every plan $U$ in $tree_{UA}$, there exists a non-empty set $\{T_1, \ldots, T_m\}$ of plans in $tree_{TO}$ such that $\mathcal{L}(U) = \{T_1, \ldots, T_m\}$.

**Proof:** It suffices to show that if plan $U_1$ is a UA-extension and plan $T_1$ is a linearization of $U_1$, then $T_1$ is a TO-extension. The proof is by induction on plan length.
*Base case*: The statement trivially holds for plans of length 0.
*Induction step*: Under the hypothesis that the statement holds for plans of length $n$, we now prove that the statement holds for plans of length $n + 1$. Suppose that $U_1$ is a UA-extension of length $n + 1$ and $T_1$ is a linearization of $U_1$. Let $U_0$ be a plan such that $U_1$ is a 1-step UA-extension of $U_0$. By the Mapping lemma, there exists a plan $T_0$ such that $T_0$ is a linearization of $U_0$ and $T_1$ is a 1-step TO-extension of $T_0$. By the induction hypothesis, $T_0$ is a TO-extension. Therefore, by definition, $T_1$ is also a TO-extension. *Q.E.D.*

**Disjointness Property:** $\mathcal{L}$ maps distinct plans in $tree_{UA}$ to disjoint sets of plans in $tree_{TO}$; that is, if $U_1, U_2 \in tree_{UA}$ and $U_1 \neq U_2$, then $\mathcal{L}(U_1) \cap \mathcal{L}(U_2) = \{\}$.

**Proof:** By the definition of $\mathcal{L}$, if $T_1, T_2 \in \mathcal{L}(U)$, then $T_1$ and $T_2$ are at the same tree depth $d$ in $tree_{TO}$; furthermore, $U$ is also at depth $d$ in $tree_{UA}$. Hence, it suffices to prove that if plans $U_1$ and $U_2$ are at depth $d$ in $tree_{UA}$ and $U_1 \neq U_2$, then $\mathcal{L}(U_1) \cap \mathcal{L}(U_2) = \{\}$.
*Base case*: The statement vacuously holds for depth 0.
*Induction step*: Under the hypothesis that the statement holds for plans at depth $n$, we prove, by contradiction, that the statement holds for plans at depth $n + 1$. Suppose that there exist two distinct plans, $U_1 = \langle \theta_1, \prec_1 \rangle$ and $U_2 = \langle \theta_2, \prec_2 \rangle$, at depth $n + 1$ in $tree_{UA}$ such that $T \in \mathcal{L}(U_1) \cap \mathcal{L}(U_2)$. Then (by definition of $\mathcal{L}$), $parent(T) \in \mathcal{L}(parent(U_1))$ and $parent(T) \in \mathcal{L}(parent(U_2))$. Since $parent(U_1) \neq parent(U_2)$ contradicts the induction hypothesis, suppose that $U_1$ and $U_2$ have the same parent $U_0$. Then, by the definition of UA either *(i)* $\theta_1 \neq \theta_2$ or *(ii)* $\theta_1 = \theta_2$ and $\prec_1 \neq \prec_2$. In the first case, since the two plans do not contain the same set of plan steps, they have disjoint linearizations and, hence, $\mathcal{L}(U_1) \cap \mathcal{L}(U_2) = \{\}$, which contradicts the supposition. In the second case, $\theta_1 = \theta_2$; hence, both plans resulted from adding plan step $O_{add}$ to the parent plan. Since $\prec_1 \neq \prec_2$, there exists a plan step $O_{int}$ that interacts with $O_{add}$ such that in one plan $O_{int}$ is ordered before $O_{add}$ and in the other plan $O_{add}$ is ordered before $O_{int}$. Thus, in either case, the linearizations of the two plans are disjoint and, hence, $\mathcal{L}(U_1) \cap \mathcal{L}(U_2) = \{\}$, which contradicts the supposition. Therefore, the statement holds for plans at depth $n + 1$. *Q.E.D.*

## A.4 Completeness Proof for TO

We now prove that TO is complete under a breadth first search control strategy. To do so, it suffices to prove that if there exists a solution to a problem, then there exists a TO-extension that is a compact solution. Before doing so, we prove the following lemma.

**Subplan Lemma:** Let totally ordered plan $T_0$ be a strict subplan of a compact solution $T_s$. Then there exists a plan $T_1$ such that $T_1$ is a subplan of $T_s$ and $T_1$ is a 1-step TO-extension of $T_0$.

**Proof:** Since $T_0$ is a strict subplan of $T_s$ and $T_s$ is a compact solution, the set of false preconditions in $T_0$, $G$, must not be empty. Let $c = \texttt{select-goal}(G)$, let $O_{need}$ be the step in $T_0$ with precondition $c$, and let $O_{add}$ be the step in $T_s$ that achieves $c$. Consider the totally ordered plan $T_1 = \langle \theta_0 \cup \{O_{add}\}, \prec_1 \rangle$, where $\prec_1 \subset \prec_s$. Clearly, $T_1$ is a subplan of $T_s$. Furthermore, by the TO-Extension Lemma, $T_1$ is a 1-step extension of $T_0$ by TO. To see this, note that $O_{add}$ is ordered before $O_{need}$ in $T_1$ since it is ordered before $O_{need}$ in $T_s$. Similarly, $O_{add}$ is ordered after the last deleter of $c$ in $T_0$ since any deleter of $c$ in $T_0$ is a deleter of $c$ in $T_s$, and $O_{add}$ is ordered after the deleters of $c$ in $T_s$. Thus, the conditions of the TO-Extension Lemma hold. *Q.E.D.*

**TO Completeness Theorem:** If plan $T_s$ is a totally ordered compact solution, then $T_s$ is a TO-extension.

**Proof:** Let $n$ be the length of $T_s$. We show that for all $k \leq n$, there exists a subplan of $T_s$ with length $k$ that is a TO-extension. This is sufficient to prove our result since any subplan of exactly length $n$ is equivalent to $T_s$. The proof is by induction on $k$.

*Base case*: If $k = 0$ the statement holds since the initial plan, which has length 0, is a subplan of any solution plan.

*Induction step*: We assume that the statement holds for $k$ and show that if $k < n$ the statement holds for $k + 1$. By the induction hypothesis, there exists a plan $T_0$ of length $k$ that is a strict subplan of $T_s$. By the Subplan Lemma, there exists a plan $T_1$ that is both a subplan of $T_s$ and a 1-step TO-extension of $T_0$. Thus, there exists a subplan of $T_s$ of length $k + 1$. *Q.E.D.*

## A.5 Completeness Proof for UA

We now prove that UA is complete under a breadth-first search strategy. The result follows from the search space correspondence defined by $\mathcal{L}$ and the fact that TO is complete. In particular, we show below that for any TO-extension $T$, there exists a UA-extension $U$ such that $T$ is a linearization of $U$. Since UA produces only unambiguous plans, it must be the case that if $T$ is a solution, $U$ is also a solution. From this, it follows immediately that UA is complete.

**Inverse Mapping Lemma:** Let $T_0 = \langle \theta_0, \prec_{t0} \rangle$ be a totally ordered plan. Let $T_1 = \langle \theta_1, \prec_{t1} \rangle$ be a 1-step TO-extension of $T_0$. Let $U_0 = \langle \theta_0, \prec_{u0} \rangle$ be a plan produced by UA such that $T_0$ is a linearization of $U_0$. Then there exists a plan $U_1$ such that $T_1$ is a linearization of $U_1$ and $U_1$ is a 1-step UA-extension of $U_0$.

**Proof:** By the definition of TO, $\theta_1 = \theta_0 \cup \{O_{add}\}$, where $O_{add}$ added some $c$ that is a false precondition of some plan step $O_{need}$ in $U_0$. Consider $U_1 = \langle \theta_1, \prec_{u1} \rangle$, where $\prec_{u1}$ is a minimal subset of $\prec_{t1}$ such that:

- $\prec_{u0} \subseteq \prec_{u1}$, and

- $(O_{add}, O_{need}) \in \prec_{u1}$, and

- $(O_{del}, O_{add}) \in \prec_{u1}$, where $O_{del}$ is the last deleter of $c$ in $U_1$, and

- no step in $U_1$ interacts with $O_{add}$

Since $\prec_{u1} \subseteq \prec_{t1}$, $T_1$ is a linearization of $U_1$. In addition, $U_1$ is an extension of $U_0$ since it meets the three conditions of the UA-Extension Lemma, as follows. First, since $c$ must have been the goal selected by TO in extending $T_0$, $c$ must likewise be selected by UA in extending $U_0$. Second, $O_{add}$ adds $c$ since $O_{add}$ achieves $c$ in $T_0$. Finally, by construction, $\prec_{u1}$ satisfies the third condition of the UA-Extension Lemma. *Q.E.D.*

**UA Completeness Theorem:** Let $T_s$ be a totally ordered compact solution. Then there exists a UA-extension $U_s$ such that $T_s$ is a linearization of $U_s$.

**Proof:** Since TO is complete, it suffices to show that if $T_1$ is a TO-extension, then there exists a UA-extension $U_1$ such that $T_1$ is a linearization of $U_1$. The proof is by induction on plan length.

*Base case*: The statement trivially holds for plans of length 0.

*Induction step*: Under the hypothesis that the statement holds for plans of length $n$, we now prove that the statement holds for plans of length $n + 1$. Assume $T_1$ is a TO-extension of length $n + 1$, and let $T_0$ be a plan such that $T_1$ is a 1-step TO-extension of $T_0$. By the induction hypothesis, there exists a UA-extension $U_0$ of length $n$ such that $T_0$ is a linearization of $U_0$. By the Inverse Mapping Lemma, there exists a plan $U_1$ that is both a linearization of $T_1$ and a 1-step UA-extension of $U_0$. Since $U_1$ is a 1-step UA-extension of $U_0$, it has length $n + 1$. *Q.E.D.*

## Acknowledgements

Most of the work present in this paper was originally described in two conference papers (Minton et al., 1991a, 1992). We thank Andy Philips for his many contributions to this project. He wrote the code for the planners and helped conduct the experiments. We also thank the three anonymous reviewers for their excellent comments.